\documentclass[letterpaper, 10 pt, conference]{ieeeconf}  

\newcommand{\ours}{SynTraC\xspace}
\IEEEoverridecommandlockouts 

\usepackage{xcolor}

\usepackage{epsfig} 
\usepackage{mathptmx} 
\usepackage{times} 
\usepackage{amsmath} 
\usepackage{amssymb}  
\usepackage{url}
\usepackage{graphicx}
\usepackage{xspace}
\usepackage{xcolor}
\usepackage{multirow}
\usepackage[capitalize]{cleveref}
\usepackage{subcaption}
\usepackage{booktabs}

\title{\LARGE \bf
SynTraC: A Synthetic Dataset for Traffic Signal Control from \\Traffic Monitoring Cameras
}

\author{Tiejin Chen$^{1*}$, Prithvi Shirke$^{1*}$, Bharatesh Chakravarthi$^{1}$, Arpitsinh Vaghela$^{1}$, Longchao Da$^{1}$, Duo Lu$^{2}$ \\ Yezhou Yang$^{1}$ and Hua Wei$^{1}$ %
\thanks{*Equal Contribution} %
\thanks{$^{1}$School of Computing and Augmented Intelligence, Arizona State University, Tempe, AZ, USA.Corresponding to {\tt\small hua.wei@asu.edu}} %
\thanks{$^{2}$Department of Computer Science and Physics, Rider University, Lawrenceville, NJ, USA. }}%

\begin{document}

\maketitle
\thispagestyle{empty}
\pagestyle{empty}

\begin{abstract}
This paper introduces \ours, the first public image-based traffic signal control dataset, aimed at bridging the gap between simulated environments and real-world traffic management challenges. Unlike traditional datasets for traffic signal control which aim to provide simplified feature vectors like vehicle counts from traffic simulators, \ours provides real-style images from the CARLA simulator with annotated features, along with traffic signal states. This image-based dataset comes with diverse real-world scenarios, including varying weather and times of day. Additionally, \ours also provides different reward values for advanced traffic signal control algorithms like reinforcement learning. Experiments with \ours demonstrate that it is still an open challenge to image-based traffic signal control methods compared with feature-based control methods, indicating our dataset can further guide the development of future algorithms. The code for this paper can be found in \url{https://github.com/DaRL-LibSignal/SynTraC}.

\end{abstract}

\section{Introduction}

Inefficient traffic signal plans contribute significantly to roadway congestion, wasting commuters' time. Traditional traffic control systems, such as the widely adopted SCATS, often depend on static, manually designed signal plans that fail to adapt to changing traffic conditions. In contrast, the advent of AI technologies and enhanced traffic monitoring capabilities, like surveillance cameras, has paved the way for innovative approaches. Recent studies have explored the application of deep reinforcement learning (RL) to traffic signal control, offering a dynamic alternative. Unlike traditional systems, RL-based methods continuously learn and adjust signal timing in response to real-time traffic data, demonstrating superior efficacy in improving traffic flow.

Despite the advances in methods, one of the critical gaps in traffic signal control (TSC) research is the absence of an image-based dataset. Existing datasets are insufficient for developing methods that interpret real-world visual cues~\cite{dai2021traffic,kunjir2022offline}. They either focus on vehicle counting without traffic signal data or use simulators that provide detailed feature data but lack real images. 
This paper addresses this gap by introducing a novel dataset that merges these two critical aspects: detailed image data and traffic signal state information. Integrating these elements into a single dataset is a significant step toward developing more sophisticated traffic management solutions that can directly leverage the visual information available from surveillance cameras. Feature-based datasets inherently lack the depth of context that visual data offers. For instance, camera data can reveal not just the number of vehicles at an intersection but also their behavior, formation, and even the presence of pedestrians or cyclists—factors that significantly influence optimal signal timing. Furthermore, with the majority of modern traffic management systems incorporating camera data for real-time monitoring and decision-making, the relevance of an image-based approach is more pronounced than ever.

This paper introduces \ours, a comprehensive dataset designed to facilitate the development of image-based TSC systems. Unlike existing datasets derived from $2D$ traffic simulators, \ours is generated using CARLA~\cite{dosovitskiy2017carla}, a sophisticated $3D$ traffic simulation platform. This approach enables collecting real-style images from roadside cameras at intersections, closely mimicking real-world conditions. Building such a dataset can also help reduce the delay of using CARLA directly. Besides Each image in \ours is accompanied by ground truth data, including bounding box locations for vehicles, making it highly compatible with computer vision technologies, such as image detection~\cite{redmon2016you}.

\begin{figure*}[t!]
\centering
\includegraphics[width=0.94\textwidth]{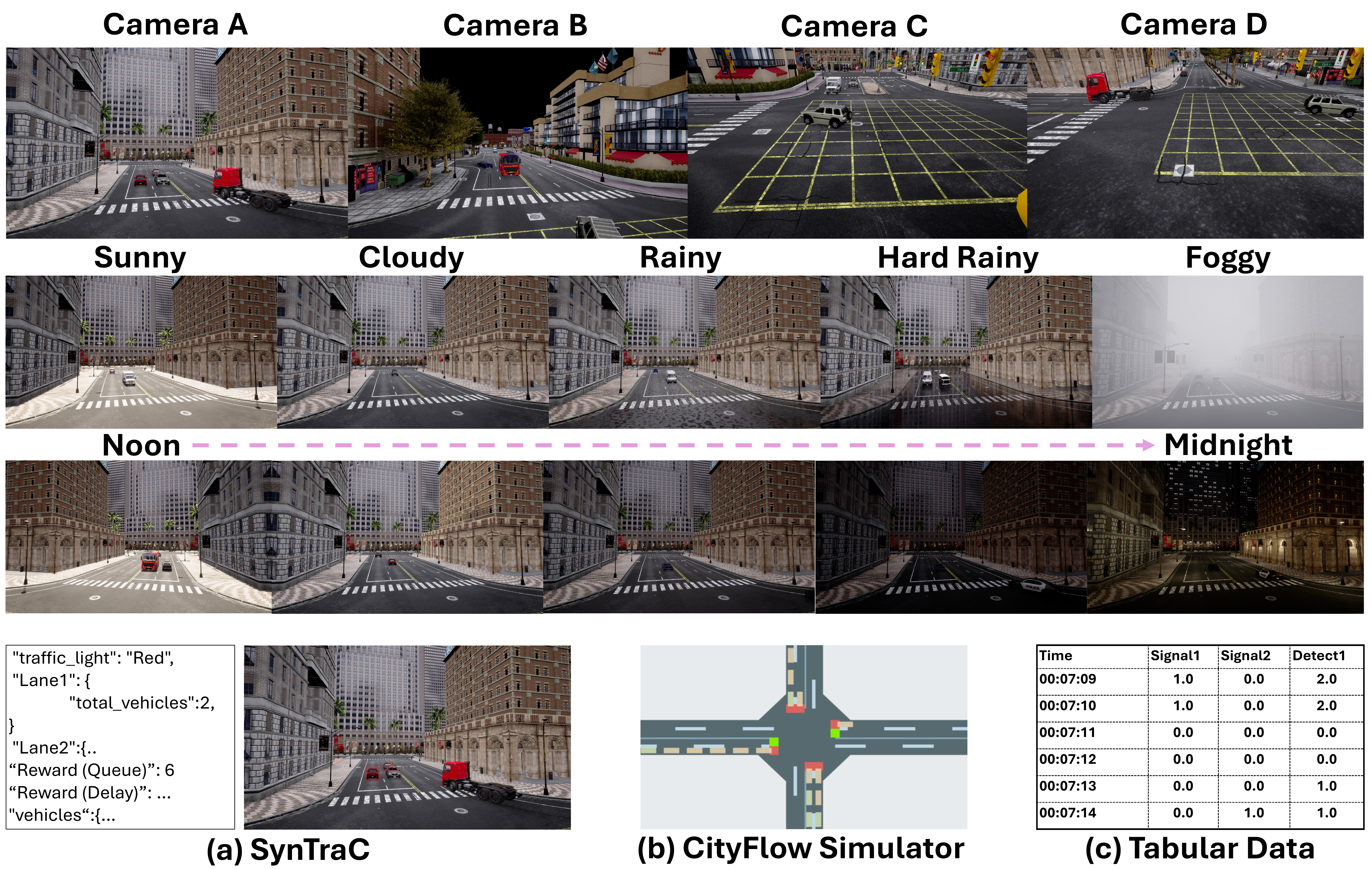}
\caption{Overview of \ours, showing the images from different cameras, weather conditions, times, and comparison with other traffic signal datasets.}
\label{fig:overview}
\vspace{-5mm}
\end{figure*}

\noindent$\bullet$~\textbf{Content and Features:}
\ours contains over 86,000 RGB images, each tagged with corresponding traffic signal states and reward values, across six hours of simulated traffic under various weather conditions and times of day. Weather conditions were diversified to encompass scenarios such as Sunny, Fog, Rainy, and Cloudy, each offering distinct challenges for traffic monitoring systems. Similarly, time conditions ranged from Day to Night. This rich dataset is further augmented with more than 250,000 reward values and approximately 225,000 vehicle bounding boxes. Additionally, we provide three rewards, i.e., queue length, waiting time, and throughput. Such detailed annotation makes \ours ideally suited for training image-based TSC policies using RL with multi-objective optimization techniques. 

\noindent$\bullet$~\textbf{Utility and Flexibility:}
Beyond its comprehensive content, \ours is designed to support advanced traffic management research and application development. To this end, we also provide the source code for dataset generation, enabling researchers and practitioners to produce customized datasets. Our automated generation pipeline offers flexibility in scenario creation, allowing for the selection of different weather conditions, times of day, and traffic flows. This adaptability ensures that users can tailor the dataset to meet specific research needs or operational challenges.

Overall, we provide the overview of \ours in \cref{fig:overview} and summarize our contributions as follows:

\noindent$~\bullet$ \ours is the first image-based dataset for TSC, which marks a foundational step towards exploring more robust strategies in traffic management. This innovation not only broadens the scope of RL applications in TSC but also sets a new benchmark for dataset complexity and relevance. \\
\noindent$~\bullet$ \ours distinguishes itself by featuring an extensive range of environmental conditions, including varied weather scenarios and times, coupled with three distinct types of reward mechanisms. This diversity offers a rich, multifaceted resource for training more robust and adaptable TSC models. \\
\noindent$~\bullet$ Comprehensive experimentation with \ours reveals a notable performance gap between image-based approaches and traditional strategies for TSC. These findings highlight the difficulties in applying existing RL methods to image-based TSC, emphasizing the need for new algorithms to fully utilize visual data in traffic management.

\section{\ours Dataset Generation Pipeline}

\begin{figure}[t]
\centering
\vspace{-4mm}
\includegraphics[width=0.45\textwidth]{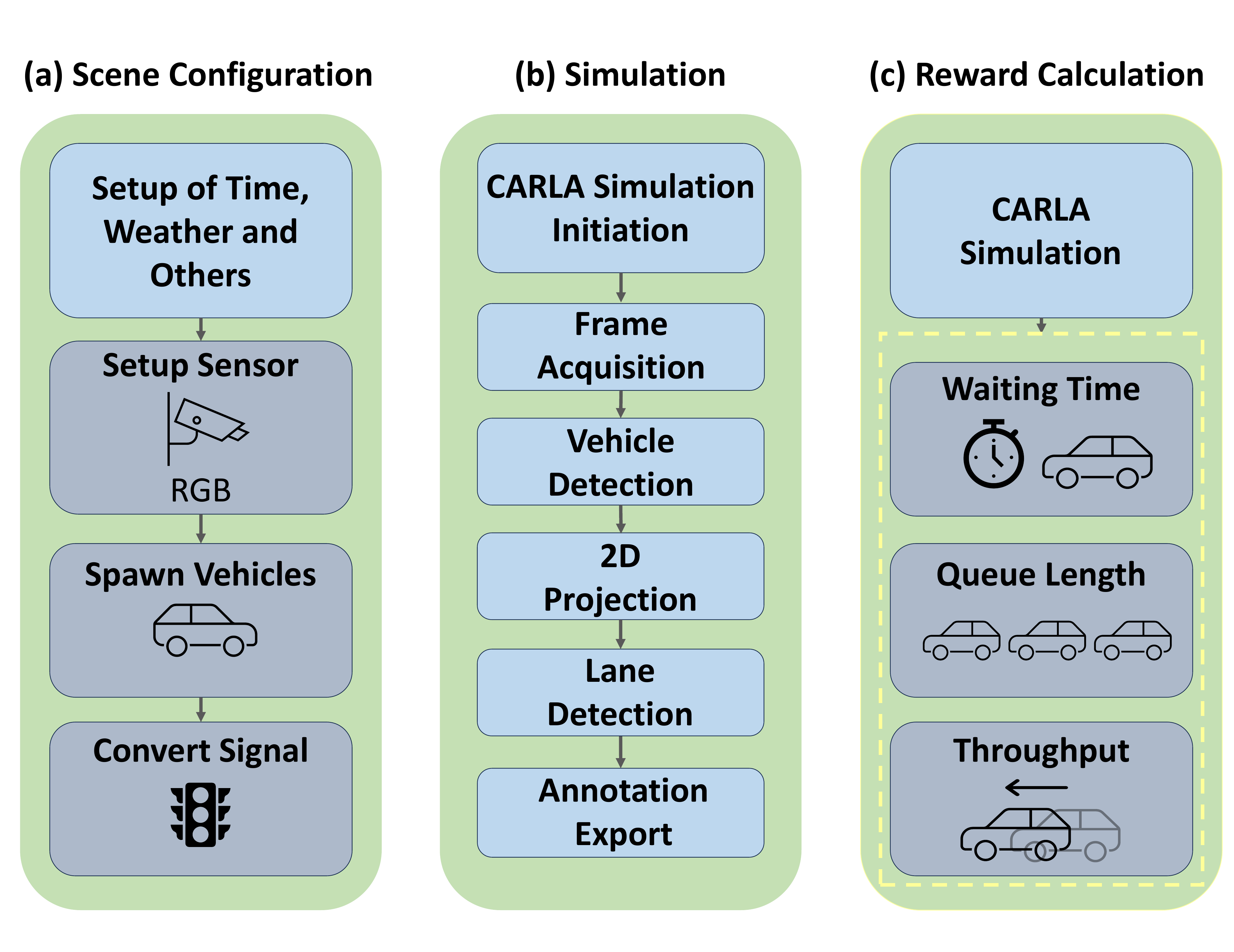}
\caption{Overview of \ours data generation pipeline.}
\label{fig:gene}
\vspace{-5mm}
\end{figure}

Our data generation pipeline is based on CARLA, a three-dimensional simulator supporting traffic signal control. However, the simulator does not have built-in support for reward calculation and lane detection which are important for creating the offline RL dataset for TSC. Hence, we developed an extension module with CARLA Python API to implement our data generation pipeline. The pipeline contains three stages, (a) traffic scene configuration, (b) simulation, and (c) reward calculation, which are illustrated in \cref{fig:gene} and detailed in the following three subsections.

\subsection{Traffic Scene Configuration}
\vspace{-2mm}
We manually set a series of traffic scenarios to ensure a diverse dataset. The environmental conditions for scenarios were set by randomizing the weather and time of the day. We chose different weather conditions such as sunny, rainy, foggy and cloudy with different time slots. 
The setup also contains 12 distinct traffic flows at an intersection with four paths, with each scenario identifying different paths that have vehicles. Such diversity of traffic flows ensures that \ours even contains various situations for RL training.

For camera setup, we set up 4 RGB cameras in the intersection in CARLA. All cameras had an image resolution of 1920$\times$1080 pixels and a field-of-view of 90$^{\circ}$. Frames were stored from the RGB camera after every second. For vehicle spawning with auto-pilot,  spawn points were chosen on the map, where vehicles were periodically spawned. Exactly one exit point was set for the map where vehicles were destroyed. Different spawn points were chosen for each traffic scenario.

Finally, we adjusted the traffic manager settings to change traffic signals with different periods of green time. Various scenarios were configured with green light times set at either 10 seconds or 15 seconds, adding variability to the traffic flow simulations. 

\subsection{Simulation and Data Generation}
We developed a Python script to gather the images and annotations during the simulation in CARLA and the gathering procedure is shown in \cref{fig:gene} (b). Firstly, we captured the image from the camera every one second. We set the CARLA server-client communication in fixed-time synchronous mode to avoid potentially skipping frames during the processing of images.

 The simulation has a fixed timestep of 0.08 seconds and CARLA takes thirteen steps (1/0.08) to recreate one second of the simulated world. We take the images of CARLA every 13 timesteps to get one image every second.

Now we could obtain images in $2D$ format from the cameras, along with information such as the traffic signal state and the speed of vehicles. However, information about coordinates in the frame, including bounding boxes and lane locations, is still in $3D$ format. Thus, we calculated the camera’s projection matrix $K$, which can project $3D$ coordinates into a $2D$ format consistent with the images. The projection matrix was derived using the focal length, which compressed the scene’s depth onto the sensor, and the principal point, which anchored the image center to the camera’s optical axis.

In detail, we have:

\begin{equation}
K = \begin{bmatrix}
\frac{S_x}{2 \cdot \tan(\frac{f \cdot \pi}{360})} & 0 & \frac{S_x}{2} \\
0 & \frac{S_x}{2 \cdot \tan(\frac{f \cdot \pi}{360})} & \frac{S_y}{2} \\
0 & 0 & 1
\end{bmatrix}
\end{equation}

Here, $S_x$ and $S_y$ are the width and the height of $2D$ images captured by cameras, respectively, and $f$ is the field of view of the cameras. $\frac{S_x}{2}$ and $\frac{S_y}{2}$ determine the principal point and $\frac{S_x}{2 \cdot \tan(\frac{f \cdot \pi}{360})}$ is the focal length. After obtaining the projection matrix $K$, we calculated the $2D$ coordinates for a given $3D$ pixel with coordinates $[x,y,z]$ using the following equation:

\begin{equation}
P = K \cdot [x,y,z]^T = [\hat{x},\hat{y},\hat{z}]^T.
\end{equation}
Subsequently, it was necessary to apply perspective projection to $\hat{x}$ and $\hat{y}$ for accounting foreshortening:
\begin{equation}
(x',y') = (\frac{\hat{x}}{\hat{z}},\frac{\hat{y}}{\hat{z}}),
\end{equation}
Where $x'$ and $y'$ are the final $2D$ coordinates.

\begin{figure}[t!]
\centering
\includegraphics[width=0.48\textwidth]{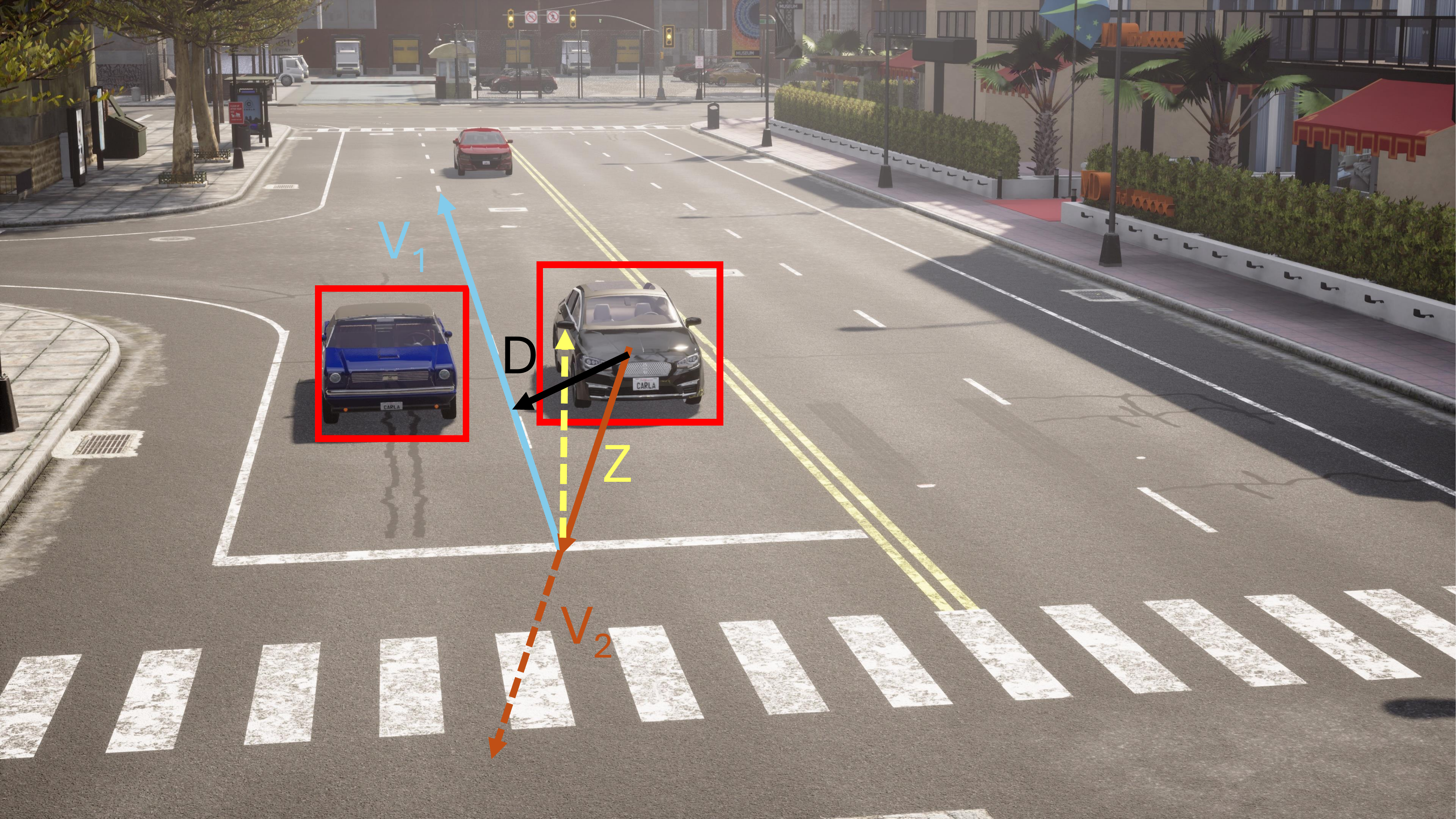}
\caption{Example of lane detection computation. The direction of cross-product $V_1 \times V_2$ points upward (yellow dash), indicating the vehicle is in the right lane.} 
\label{fig:lane_detection}
\vspace{-6mm}
\end{figure}

To fulfill the TSC's requirement for counting vehicles within each lane, we conducted lane detection on the $2D$ images. The overall calculation is shown in \cref{fig:lane_detection}. The start and end points of the lane are predetermined by CARLA and projected into $2D$ images through the projection matrix $K$. Next, $V_1$, the lane reference vector connecting the lane’s start and end points, and $V_2$, extending from the vehicle’s bounding box center to the lane’s start point, were obtained and shown as the blue and orange lines in \cref{fig:lane_detection}, respectively.

The distance between the lane vector and the vehicle's center, denoted as $D$ and illustrated in black in \cref{fig:lane_detection}, acts as a threshold to define the search region where only vehicles in the search region can be detected.

By computing the cross product $z = V_1 \times V_2$ and using the right-hand rule to determine its direction, we could ascertain the vehicle's lane orientation. The vector $z$, shown as a yellow dash in \cref{fig:lane_detection}, points upward for vehicles in the right lane and downward (indicating a negative cross-product result) for those in the left lane as observed from the cameras.

\subsection{Reward Calculation}
We calculated three different rewards related to TSC for \ours during generation.  The reward is an RL term that measures the return by taking action under the state. Considering a diversity of rewards, our dataset contains:
\\\noindent$~\bullet$ Waiting Time (WT) measures the amount of time that vehicles spend waiting in the network. For a control policy, the smaller WT indicates a better quality of TSC.
\\\noindent$~\bullet$ Queue length is the number of vehicles waiting to pass through the intersection in the road network. A good TSC policy should not make too many vehicles wait. Therefore, a smaller queue length indicates a better quality of TSC.
\\\noindent$~\bullet$ Throughput (TP) is the number of vehicles that reach their destinations within an amount of time. A larger TP, which means more vehicles are arriving at their destinations, indicates a better traffic flow and thus a better TSC.

Queue length was calculated for each camera individually when we took the image every time. Waiting Time was also calculated for each camera and we summed up the waiting time from every stopped vehicle which has a speed of less than 0.1m/s. 
In contrast, throughput was calculated as a sum of all the cameras together, and the value of throughput is accumulated through the simulation and never decreases unless we start a new simulation for a new traffic scene.

\section{EXPERIMENTAL EVALUATION}

Our experiments focus on evaluating the image-based traffic signal control methods trained with our image dataset on both synthetic and real-world data. Every image-based traffic signal control method contains a detection model and count-based control policy. 

For the online testing, we adopt the CARLA and several types of traffic flow in CARLA to make a variant test environment. 
The whole simulation duration is 240 seconds with 3 different traffic flows. Consistent with the previous count-based TSC methods, we first obtain the control policies with ground truth counts for each lane and then acquire the counts for each lane through image detection models. We do not directly use an end-to-end training method using images as input because the resolution of images is so high that training time will become unacceptable. The random seed is set to the same for every experiment.

\subsection{Evaluation Setting}
We adopt the commonly used metrics, detection models, and RL models for our evaluation.

\textbf{RL Model.} 
We adopt four different popular RL training methods that fit offline RL training: 1) Deep Q-Network~(DQN)~\cite{mnih2013playing}; 2) Double DQN (DDQN)~\cite{van2016deep}; 3) Soft Actor-Critic~(SAC)~\cite{haarnoja2018soft}; and 4) Conservative Q-Learning~(CQL)~\cite{kumar2020conservative}. We also compared RL-based control policies with traditional methods such as MaxPressure, which is a rule-based method using count information as well and we use ground truth counts for MaxPressure.

\textbf{Detection Models.} Our final TSC methods contain
the image detection models to get the vehicle numbers for each lane. To test the influence of different detection models, we consider 3 detection models: 1) Masked R-CNN~\cite{he2017mask}; 2) Faster R-CNN~\cite{girshick2015fast} and 3) RetinaNet~\cite{lin2017focal}. We mainly focus on the pre-trained detection models while we also provide the results for fine-tuned detection models on \ours.

\textbf{Metrics.} For evaluating the performance of image detection models, we mainly use Mean Square Error~(MSE) and Mean Absolute Error~(MAE) to evaluate the detected vehicle numbers with ground truth vehicle numbers. For evaluating control policies, we use total WT~(in seconds), queue length (in counts), and throughput which we introduced in the previous section. We also consider travel time~(TT) which calculates how much time a vehicle requires to arrive at the destination. 

\subsection{Evaluation Results}
\begin{table}[th!]
\centering
\renewcommand{\arraystretch}{0.8}
\begin{tabular}{cccc}
\toprule
\textbf{Visual Models}& \textbf{MSE}($\downarrow$)  & \textbf{MAE}($\downarrow$) & \textbf{Inference Time}($\downarrow$)\\
\midrule
\multicolumn{4}{c}{Sunny--Day Time} \\
\midrule
Faster R-CNN & \textbf{1.01} & \textbf{0.64} & \textbf{9.31} \\
Masked R-CNN & 1.03 & 0.64 & 11.60  \\
RetinaNet & 1.07 & 0.66 & 9.49 \\
\midrule
\multicolumn{4}{c}{Clear--Night Time} \\
\midrule
Faster R-CNN & \textbf{1.31} & \textbf{0.74} & \textbf{8.64} \\
Masked R-CNN & 1.26& 0.73 & 9.72\\
RetinaNet & 2.29 & 1.02 & 8.71 \\
\midrule
\multicolumn{4}{c}{Rainy--Day Time} \\
\midrule
Faster R-CNN & \textbf{1.01} & \textbf{0.65} & 9.53 \\
Masked R-CNN &1.04 & 0.66 & 10.96 \\
RetinaNet & 1.18 & 0.78 & \textbf{9.46} \\
\midrule
\multicolumn{4}{c}{Fog--Day Time} \\
\midrule
Faster R-CNN & 2.10 & 1.02 & \textbf{8.84} \\
Masked R-CNN & \textbf{1.87} & \textbf{0.95} & 10.06 \\
RetinaNet &2.47 & 1.13 & 8.87 \\
\bottomrule
\end{tabular}
\caption{Performance of image detection models under different weather and times. Performances drop in other weather and times compared with sunny and daytime.}
\label{tab:count_eva}
\vspace{-5mm}
\end{table}

\begin{table*}[th]
\centering
\resizebox{0.95\textwidth}{!}{\begin{tabular}{cccccccccccccccccccc}
\toprule
\multirow{2}{*}{\textbf{Models}} & \multicolumn{4}{c}{\textbf{Sunny--DayTime}} & & \multicolumn{4}{c}{\textbf{Rainy--DayTime}}& & \multicolumn{4}{c}{\textbf{Clear--NightTime}} & & \multicolumn{4}{c}{\textbf{Rainy--NightTime}}\\ 
\cmidrule{2-5}  \cmidrule{7-10}  \cmidrule{12-15} \cmidrule{17-20}
 & \textbf{WT}($\downarrow$) & \textbf{QL}($\downarrow$) & \textbf{TT}($\downarrow$) & \textbf{TP}($\uparrow$) & & \textbf{WT} ($\downarrow$) & \textbf{QL} ($\downarrow$) & \textbf{TT} ($\downarrow$) & \textbf{TP} ($\uparrow$)& & \textbf{WT} ($\downarrow$) & \textbf{QL} ($\downarrow$) & \textbf{TT}($\downarrow$) & \textbf{TP}($\uparrow$) & & \textbf{WT}($\downarrow$) & \textbf{QL} ($\downarrow$) & \textbf{TT}($\downarrow$) & \textbf{TP} ($\uparrow$)\\ \midrule
Fix-Time & 3318.14 & 223 & 42.98 & 82 & & 3318.14 & 223 & 42.98 & 82 & & 3318.14 & 223 & 42.98 & 82 & & 3318.14 & 223 & 42.98 & 82 \\
MaxPressure & 3309.78 & 183 & 32.68 & 95 & & 3309.78 & 183 & 32.68 & 95 & & 23309.78 & 183 & 32.68 & 95 & & 3309.78 & 183 & 32.68 & 95
\\ \midrule
\multicolumn{20}{c}{Ground Truth Count} \\ \midrule
DQN & 2600.21 & 172 & 26.39 & 106 & & 2600.21 & 172 & 26.39 & 106 & & 2600.21 & 172 & 26.39 & 106 & & 2600.21 & 172 & 26.39 & 106  \\
DDQN & 3289.17 & 200 & 39.03 & 93 & & 3289.17 & 200 & 39.03 & 93 & & 3289.17 & 200 & 39.03 & 93 & & 3289.17 & 200 & 39.03 & 93 \\
SAC & 2474.60 & 170 & 24.86 & 108 & & 2474.60 & 170 & 24.86 & 108 & & 2474.60 & 170 & 24.86 & 108 & & 2474.60 & 170 & 24.86 & 108 \\
CQL & \underline{1439.84} & \underline{123} & \underline{23.73} & \underline{112} & & \underline{1439.84} & \underline{123} & \underline{23.73} & \underline{112} & & \underline{1439.84} & \underline{123} & \underline{23.73} & \underline{112} & & \underline{1439.84} & \underline{123} & \underline{23.73} & \underline{112} \\ 
\midrule
\multicolumn{20}{c}{Faster R-CNN} \\
\midrule
DQN & 2570.24 & 196 & \textbf{28.51} & \textbf{102} & & \textbf{2644.63} & 206 & \textbf{26.04} & \textbf{104} & & 3896.33 & 272 & 62.40 & 55 & & 3002.63 & 255 & 59.21 & \textbf{73} \\ 
DDQN & 3536.71 & 265 & 33.47 & 88 & & 2852.42 & \textbf{202} & 36.14 & 97 & & 3911 & 290 & 57.41 & 62 & & \textbf{2653.16} & \textbf{221} & 50.67 & 63 \\
SAC & 4457.22 & 280 & 47.03 & 87 & & 3528.63 & 234 & 42.58 & 96 & & 3577.12 & 278 & \textbf{48.07} & \textbf{57} & & 3187.34 & 250 &  \textbf{41.31} & 68\\
CQL & \textbf{1918.24} & \textbf{156} & 29.61 & 98 & & 3338.51 & 223 & 31.13 & 97 & & \textbf{2768.57} & \textbf{216} & 61.74 & 60 & & 3005.56 & 227 & 60.06 & 57\\ \midrule
\multicolumn{20}{c}{RetinaNet} \\ \midrule
DQN & \textbf{2030.72} & \textbf{157} & 30.28 & 93 & & 3663.18 & 227 & 48.88 & 81 & & 3343.57 & \textbf{237} & 57.32 & 43 & & 4278.94 & 329 & 59.51 & 51 \\
DDQN & 3120.43 & 209 & 43.76 & 85 & & 3051.15 & \textbf{214} & 38.46 & 92 & & \textbf{3113.86} & 256 & 53.71 & \textbf{57} & & 2986.20 & 219 & 54.12 & \textbf{62}\\
SAC & 3038.81 & 228 & \textbf{28.35} & \textbf{99} & & \textbf{2717.88} & 244 & \textbf{28.53} & \textbf{102} & & 4187.30 & 281 & \textbf{45.46} & 42 & & \textbf{2951.77} & \textbf{196}& 49.35 & 56\\
CQL & 3449.66 & 285 & 33.26 & 55 & & 3390.01 & 234 & 47.13 & 79 & & 3559.40 & 284 & 61.53 & 46 & &3271.10 & 345 & \textbf{39.05} & 60\\ \bottomrule
\end{tabular}}
\caption{Performance of different methods under different weather and times. Every testing method contains an image detection model (Faster R-CNN or RetinaNet) and a control model (Fix-Time, DQN, DDQN, SAC, or CQL). The results show that RL-based methods have a better performance compared with Fix-Time. Performances drop when image detection models are used, especially at night. We \underline{underline} top results without detection models and \textbf{highlight} the best with them.}
\label{tab:day_time_res}
\vspace{-7mm}
\end{table*}

We use three different pre-trained detection models and four offline RL models. We evaluate the performance of different detection models and the image-based TSC methods.

\textbf{Evaluation of Detection Models.} In \cref{tab:count_eva}, we present the results of evaluating image detection models. We have the following main observations:
\\\noindent$~\bullet$ Compared to other methods, faster R-CNN has a better result across different weather and times. This is possible because other more advanced methods are overfitting to the pre-trained dataset. Besides, faster R-CNN has a better inference time which is another advantage in the real world.
\\\noindent$~\bullet$ Different weather and times influence the performance of all detection models. Compared with performance under sunny and daytime, the performance of all models drops in other weather and times. We find that nighttime influences the performance the most because most images in the pre-trained dataset are taken in well-lit conditions to depict the objects. The variety of images in \ours plays an important role in fine-tuning detection models that be used at night.

\textbf{Evaluation of Control Policies.} In the \cref{tab:day_time_res}, we present the results of evaluating control models (with detection models together) and we have the following observations:\\
\noindent$~\bullet$ Compared with the rule-based policies, all RL-based models using ground-truth counts in each lane are better in all metrics except for DDQN, showing the superiority of RL-based TSC. This observation is consistent with previous works~\cite{wei2018intellilight,zheng2019learning}.
\\\noindent$~\bullet$ Compared with using ground truth count values, after combining the detection models, the performance drops a lot even with sunny and daytime when detection models have the best performance. This indicates that improving the detection models or proposing new algorithms handling the non-accurate count information is required in the feature. 
\\\noindent$~\bullet$ Among all RL models, CQL works the best in all metrics, showing that CQL has the advantage of utilizing the limited data setting in offline reinforcement learning. However, CQL also drops the performance the most when using detection models. This is reasonable since a better policy is more sensitive to the wrong information.
\\\noindent$~\bullet$ When the weather becomes rainy, the drop in performance is insignificant compared with sunny. When time becomes nighttime, the performance drop is more significant, which is consistent with the detection model's performance. 

\subsection{Influence of Different Camera Angles} We evaluate our methods with three different camera angles in CARLA. We provide the evaluation results for raising and lowering the cameras' angles and raising the location of cameras to provide a bird-view image in \cref{tab:camera_angle}. 
\\\noindent$~\bullet$ Changing the angles of cameras will influence the RL policies significantly. After raising the cameras, the performance of DQN becomes better even compared with using ground truth count values. This is because DQN performs better if the number of vehicles decreases. Raising the cameras' angles will cause the detection numbers to become less. Therefore, the performance of DQN is even better than using ground truth count information. The same thing happens when we lower the cameras. The performance increases especially for SAC because lowering the cameras can output the detection results that fit better for SAC.
\\\noindent$~\bullet$ Raising the locations of cameras captures images having the most comprehensive information to the detection model and thus the performance of CQL is the best in this case. Besides, the performances of SAC and DQN become similar to using the ground truth count information.

\begin{table}[th]
\centering
\renewcommand{\arraystretch}{0.8}
\setlength\tabcolsep{5pt}
\vspace{-3mm}
\begin{tabular}{cccc}
\toprule
\textbf{RL Models}& \textbf{Waiting Time}($\downarrow$) & \textbf{Queue Length}($\downarrow$)  & \textbf{Throughput} ($\uparrow$)\\
\midrule
\multicolumn{4}{c}{Raise the Angles of Cameras} \\
\midrule
DQN & \textbf{1223.38}&\textbf{125} & \textbf{106}\\
SAC & 2266.11&172 & 95\\
CQL &2523.95 &188  &104 \\
\midrule
\multicolumn{4}{c}{Lower the Angles of Cameras} \\
\midrule
DQN &1658.16 & 141  & 99 \\
SAC & \textbf{1262.34} & \textbf{132} & \textbf{109} \\
CQL & 1655.32& 180 & 101\\
\midrule
\multicolumn{4}{c}{Raise the Location of Cameras} \\
\midrule
DQN & 2756.81&190 & 90\\
SAC & 2246.08 & 174& 102\\
CQL & \textbf{1416.39}& \textbf{144} & \textbf{113} \\
\bottomrule
\end{tabular}
\caption{Performance of different methods under different camera angles. All methods are not robust to camera angles.}
\label{tab:camera_angle}
\vspace{-5mm}
\end{table}

\subsection{Generalization on Real-world Data} 
We evaluate the image detection models and TSC model learned with SynTraC dataset using a real-world dataset from an intersection at Tempe, Arziona. By doing so, we justify our dataset uniformly enhances a better generalizability on both image detection methods and prevalent end-to-end image-based TSC methods.

\textbf{Evaluation on Detection Model.} First, we evaluate the detection models fine-tuned with \ours by comparing them with the per-trained detection model. We use RetinaNet as the detection method and annotate 100 images from a 6-minute video captured by the cameras from the intersection with the count of vehicles. The comparison results are provided in \cref{tab:real_count}. From the results, we can see that fine-tuning with \ours increases the performance of the detection model even in the real world, \textbf{indicating that a generalization ability of \ours in the real world and \ours can help to develop better image-based TSC methods in the real world.}

\begin{table}[th!]
\centering
\begin{tabular}{cccc}
\toprule
\textbf{Visual Models}& \textbf{MSE}($\downarrow$)  & \textbf{MAE}($\downarrow$) & \textbf{Inference Time}($\downarrow$)\\
\midrule
Pre-trained Model & 1.975 & 0.975 & \textbf{9.42} \\
Fine-tuned Model& \textbf{1.725} & \textbf{0.925} & 9.43 \\
\bottomrule
\end{tabular}
\caption{Comparison of pre-trained image detection model and fine-tuned model using \ours on the real-world data. Fine-tuned Model generalizes in the real world.}
\label{tab:real_count}
\vspace{-4mm}
\end{table}

\begin{figure}[h!]
\centering
\includegraphics[width=0.48\textwidth]{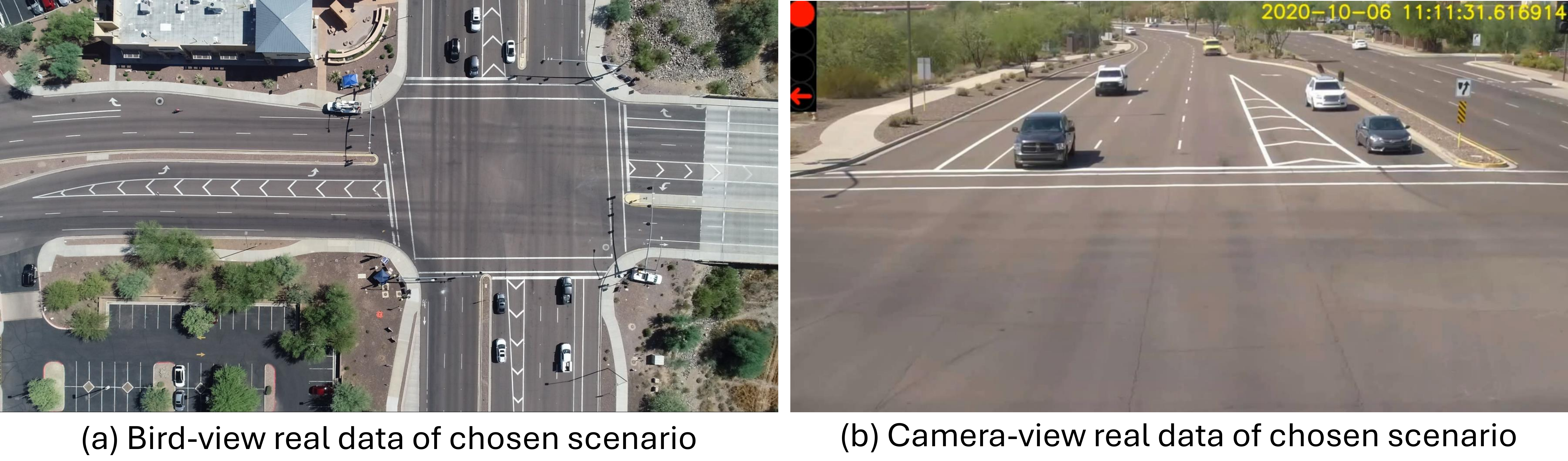}
\vspace{-5mm}
\caption{Example images for our chosen scenario. In the bird-view image, vehicles wait north and south while no east or west car coming. The traffic signal is red in this scene.} 
\label{fig:real}
\vspace{-3mm}
\end{figure}
\textbf{Evaluation on TSC Methods.} Since we cannot directly change the traffic signal states on the real-world data we collected, we do not provide a detailed evaluation of normal metrics such as waiting time. To evaluate image-based TSC methods using real-world data, we utilize scenarios where the real-world TSC method exhibited suboptimal performance. In this scenario, we applied our end-to-end TSC method, trained with \ours (utilizing detection models and the control policy trained with \ours), to obtain the decision. This approach allows us to demonstrate the effectiveness of our method in addressing real-world challenges. In detail, we can get one real-world scenario shown in \cref{fig:real}. From the images, we can see that the vehicles are waiting in the direction of north and south for the traffic signal state to become green while no cars coming from west or east. In fact, in this scenario, the red light continues for around 30 seconds and more than 10 vehicles are waiting. In contrast, only 3 cars from west or east go through this intersection in this period. Therefore, from a human perspective, it is a bad decision made by real-world control. However, when we feed the same scenario into the end-to-end traffic signal control method trained with \ours, our method outputs it should change the traffic signal states, which is a correct decision.

\section{Discussion}

Recent datasets~\cite{genser2023traffic,van2019open} on traffic signal control offer traffic signal timings with traffic states. However, none of these datasets are suitable for RL approaches, which have gained much attention in recent traffic signal control area~\cite{wei2018intellilight,wei2019presslight} since they only contain information from traffic signals, which is less informative for RL approaches. Most current RL approaches are based on traffic simulators like Cityflow~\cite{zhang2019cityflow,da2024cityflower} where RL policy takes the counts of vehicles in each lane as the inputs for online training. There also exists some offline RL datasets. For example,~\cite{zhang2023data} collected data from Cityflow for offline RL while~\cite{dai2021traffic,kunjir2022offline} collected data from SUMO simulator~\cite{SUMO2018,mei2023libsignal}. However, all of these simulators or datasets do not contain traffic monitoring images~\cite{jain2019review} considering the difficulty of directly getting the count information from real-world sensors. We need to utilize more realistic data such as traffic monitoring images. Our dataset \ours fills the blank in this area.

\section{Conclusion and Future Work}

In this paper, we introduce \ours, the first image-based dataset with reward data for traffic signal control (TSC) using offline reinforcement learning, along with a generation pipeline that aligns TSC with real-world conditions. Besides, \ours also contains information for other downstream tasks such as vehicle detection. The evaluation on \ours shows that the RL policies that control the traffic signal can perform very well using ground truth information while integrating the detection models into the TSC system hurts the performance of RL policies.

In the future, We aim to propose new control algorithms designed to handle uncertain information, thereby preventing performance degradation of the detection model under various weather conditions. Besides, a thorough evaluation of the end-to-end TSC system trained with \ours in the real world is also in our next plan.

\section*{Acknowledgements}
The work was partially supported by NSF awards \#2421839. The views and conclusions contained in this paper are those of the authors and should not be interpreted as representing any funding agencies.

\bibliographystyle{plain}
\bibliography{IEEEfull}

\end{document}